\documentclass[10pt,twocolumn,letterpaper]{article}

\usepackage{cvpr}
\usepackage{times}
\usepackage{epsfig}
\usepackage{graphicx}
\usepackage{amsmath}
\usepackage{amssymb}
\usepackage{authblk}
\usepackage{color}
\usepackage{multirow}
\usepackage{tabularx}

\newcolumntype{Y}{>{\centering\arraybackslash}X}

\usepackage[breaklinks=true,letterpaper=true,colorlinks,bookmarks=false]{hyperref}

\cvprfinalcopy 

\def\cvprPaperID{2} 

\ifcvprfinal\fi
\begin{document}

\title{CUHK \& ETHZ \& SIAT Submission to ActivityNet Challenge 2016}

\author[1]{Yuanjun Xiong} 
\author[2]{Limin Wang}
\author[3]{Zhe Wang}
\author[3]{Bowen Zhang}
\author[1]{Hang Song}
\author[1]{Wei Li}
\author[1]{Dahua Lin}
\author[3]{Yu Qiao}
\author[2]{Luc Van Gool}
\author[1]{Xiaoou Tang}     
\affil[1]{Multimedia Laboratory, The Chinese University of Hong Kong, Hong Kong}
\affil[2]{Computer Vision Lab, ETH Zurich, Switzerland}
\affil[3]{Shenzhen Institutes of Advanced Technology, CAS, China} 

\renewcommand\Authands{ and }

\maketitle

\begin{abstract}
This paper presents the method that underlies our submission to the untrimmed video classification task of ActivityNet Challenge 2016. We follow the basic pipeline of temporal segment networks~\cite{Wang2016ECCV} and further raise the performance via a number of other techniques. Specifically, we use the latest deep model architecture, e.g., ResNet and Inception V3, and introduce new aggregation schemes (top-k and attention-weighted pooling). Additionally, we incorporate the audio as a complementary channel, extracting relevant information via a CNN applied to the spectrograms. With these techniques, we derive an ensemble of deep models, which, together, attains a high classification accuracy (mAP $93.23\%$) on the testing set and secured the first place in the challenge.  
\end{abstract}

\begin{table*}[!ht]
	\centering
	\caption{Performance of different network architectures on ActivityNet v1.3 validation set. Performance is measured by per-class mean average precision (mAP) and top-$3$ prediction accuracy. We use the variant ``basic+a'' in training these models.}
	\label{tb:cnn}
	\begin{tabularx}{1\textwidth}{|c *{6}{|Y}|}
		\hline
		\multirow{2}{*}{Settings} & \multicolumn{3}{c|}{Spatial Nets}       & \multicolumn{3}{c|}{Temporal Nets}   \\ \cline{2-7} 
		& BN-Inception & Inception V3 & ResNet    & BN-Inception & Inception V3 & ResNet \\ \hline
		mAP                       & $79.7\%$    & $83.3\%$    & $83.3\%$ & $63.3\%$    & $64.3\%$    & -      \\ \hline
		Top-3 Acc.            & $89.6\%$    & $91.5\%$    & $91.6\%$ & $77.0\%$    & $77.9\%$    & -      \\ 
		\hline
	\end{tabularx}
\end{table*}

\section{Introduction}

In the past several years, the advance in deep learning techniques has given rise to a new wave of efforts towards vision-based action understanding.
A number of deep learning based frameworks, including two-stream CNNs~\cite{SimonyanZ14}, 3D CNNs (C3D)~\cite{Du2015C3D}, and Trajectory-pooled Deep convolutional Descriptors (TDD)~\cite{WangQT15a},
have been developed, which significantly pushed forward the state-of-the-art~\cite{WangS13a, WangQT16}.
Such improvement on performance, to a large extent, is owning to both the modeling capacity of deep architectures and more effective learning strategies.

However, it is worth noting that previous efforts focus mainly on the analysis of short video clips. These clips are typically extracted from longer videos such that they only contain the portions of frames that truly capture the actions of interest. Obviously, preparation of such data is a laborious procedure. 
Action recognition from \emph{untrimmed videos}, a problem that is more pertinent to real-world demands, is drawing increasing attention from the community. While substantially reducing the efforts needed in manual annotation, this task on the other hand presents a new challenge to the recognition system -- a significant (or even dominant) fraction of a given video is irrelevant to the action of interest.

Driven by the ActivityNet benchmark~\cite{Caba2015anet}, we develop an integrated approach to recognizing actions from untrimmed videos\footnote{Codes and models are available  at \url{https://github.com/yjxiong/anet2016-cuhk}}. 
Our approach follows the framework of temporal segment networks presented in our earlier paper~\cite{Wang2016ECCV}, which allows modeling long-range temporal structure in actions and introduces various techniques to improve the training procedure,  \textit{e.g.}~temporal pre-training, and scale jittering augmentation. On top of this framework, we develop several new techniques to further improve the recognition accuracy.
While visual analysis plays a primary role in this task, we notice that the audio channels that come with these videos provide complementary information. To exploit such information, we develop a deep network called Audio CNN to derive complementary features from the spectrograms. 

Combining both the visual and acoustic models, we attain a high recognition accuracy (mAP $93.23\%$ on testing set). We want to emphasize that this performance is obtained only using the training data provided by the ActivityNet benchmark except using CNNs pre-trained on ILSVRC12 data for initialization -- no additional data or annotations are used throughout both the training and testing procedures. 

The rest of this paper is organized as follows. Section 2 presents our approach in detail, Section 3 reports our results under a variety of settings, finally Section 4 concludes this work.

\section{Our Approach}

Our approach to  untrimmed video classification comprises two complementary components: visual and acoustic modeling.
The visual analysis, which combines a variety of techniques, plays a primary role in this framework, while the acoustic model exploits complementary information from the audio channels to further improve the performance. Next, we present these components respectively in Section 2.1 and 2.2.

\subsection{Visual Analysis System}

Our visual analysis component works as follows: it samples multiple snippets from a given video, makes snippet-wise predictions using very deep two-stream CNNs, and finally aggregates the predictions via different strategies such as top-k and attention-weighted pooling.

\paragraph{Snippet-wise Predictor}\label{sec:snippet}
Deep convolutional neural networks (CNN) which learns from multiple modality of input data has been used extensively in visual recognition tasks~\cite{SzegedyLJSRAEVR14, Xiong2015Event, Zhang2016MV,Gan15} and achieved superiority over models using a single modality.
The snippet-wise predictor in our approach is a realization of temporal segment network framework~\cite{Wang2016ECCV} which consists appearance and motion modeling parts.
In this work, we adopt the recently proposed network architectures such as \textbf{ResNet}~\cite{He2015ResNet} and \textbf{Inception V3}~\cite{Szegedy2015ICP2} to improve the capacity of the frame-wise predictor.

During training of the snippet-wise predictor, the techniques introduced in~\cite{Wang2016ECCV}, such as scale jittering and stronger dropout, are also applied to the these architectures.
The basic idea of temporal segment networks is to sample several snippets from one input video to jointly train the CNNs by averaging the per-snippet prediction.
We also experimented with more advanced aggregation techniques into the training process.

\paragraph{Video-level Classification}
To obtain video-level classification results, we use the following strategy:
the snippet-wise predictor is first applied to an input video snippet with a $1$FPS sampling rate, then an aggregation module will combine the snippet-wise class scores into the final prediction. 
We experimented with several advanced strategies for combing snippet-wise scores of the appearance nets.
These include top-$k$ pooling and attention weighted pooling.
These strategies, when used in both training and testing, produced models that are complementary to each other and thus form effective components in the final ensemble.

\begin{table}[t]
	\centering
	\caption{Performance comparison of the appearance modeling CNN variants on the validation set of ActivityNet v$1.3$. Here we analyze their performance using the Inception V3~\cite{Szegedy2015ICP2} architecture.
		In the table, ``basic'' refers to the baseline approach in~\cite{Wang2016ECCV}, ``a'' refers to models trained with multiple snippets from one video, ``b'' refers to models equipped with advanced aggregation strategies.
	}
	\vspace{5pt}
	\label{tb:cnn_tricks}
	\begin{tabularx}{1\linewidth}{|l *{2}{|Y}|}
		\hline
		Variants     	& mAp      & Top-3 Acc. \\ \hline
		basic      		& $82.9\%$ & $91.0\%$   \\ \hline
		basic+a         & $83.3\%$ & $91.5\%$   \\ \hline
		basic+ab        & $84.2\%$ & $92.1\%$   \\ \hline
		Ensemble 		& $85.9\%$ & $92.9\%$   \\ \hline
	\end{tabularx}
\end{table}

\begin{table}[t]
	\centering
	\caption{Performance of different components in the visual analysis system on the validation set.
		Here, ``Appearance CNN'' refers to the appearance modeling part. ``Motion CNN'' refers to the motion modeling part. ``Combined CNN'' refers to the results by combining both  appearance and motion modeling parts.
		``Visual All'' refers to the results by further combining scores from other methods such as IDT~\cite{WangS13a,PengWWQ14} and TDD~\cite{WangQT15a}.
	}
	\vspace{5pt}
	\label{tb:visual}
	\begin{tabularx}{1\linewidth}{|l *{2}{|Y}|}
		\hline
		Variants     		& mAp      & Top-3 Acc. \\ \hline
		Appearance CNN    	& $85.9\%$ & $92.9\%$   \\ \hline
		Motion CNN        	& $68.3\%$ & $80.2\%$   \\ \hline
		Combined CNN     	& $89.7\%$ & $95.0\%$   \\ \hline
		Visual All 			& $90.4\%$ & $95.2\%$   \\ \hline
	\end{tabularx}
\end{table}

\subsection{Acoustic Analysis System}
Audio signals in a video carry important cues for recognizing some action classes. 
To harness the information in this aspect, we combine the standard MFCC~\cite{O2008MFCC} representations with audio-based CNNs~\cite{Naoya2016Audio,WuJWYXW15} to form the acoustic modeling system.

\paragraph{MFCC}
Mel Frequency Cepstral Coefficients (MFCC) ~\cite{O2008MFCC} is a powerful feature descriptor used in automatic speech recognition system.
In our approach, we extract MFCC features from companioned audios of the videos in the dataset, and train SVMs on descriptors aggregated with Fisher Vector~\cite{SanchezPMV13}

\paragraph{Audio CNN}
The basic idea of Audio CNN works is to apply CNNs on spectrograms, or time-frequency-response maps,  of audio signals.
In this work, we propose to directly use the \emph{grayscale} time-frequency map image to train the audio CNN.
Then the audio CNN can be initialized by the same technique used on the temporal networks in~\cite{Wang2016ECCV}.
It is also known that learning from multiple time scales help in acoustic models~\cite{Zhu2016Wave}.
In this sense, we propose to stack multiple spectrograms with varying window size as the input to the audio CNN.

\section{Experiments}

\begin{table}[t]
	\centering
	\caption{Performance of acoustic models on ActivityNet v1.3 validation set. Performance is measured by per-class mean average precision (mAP) and top-$3$ prediction accuracy.
		Here, ``Gray'' refers to the models trained with grayscale inputs. ``MS'' refers to the model trained with multiple time scales.}
	\vspace{5pt}
	\label{tb:audio}
	\begin{tabularx}{1\linewidth}{|l *{2}{|Y}|}
		\hline
		Methods           & mAP & Top-3 Acc. \\ \hline
		MFCC (FV+SVM)     & $14.2\%$    &  $26.1\%$          \\ \hline
		Audio CNN         & $8.0\%$    &  $17.1\%$          \\ \hline
		Audio CNN Gray    & $9.3\%$    &  $19.3\%$          \\ \hline
		Audio CNN Gray+MS & $10.3\%$    & $20.7\%$          \\ \hline
		Audio Ensemble    & $15.2\%$    & $29.1\%$          \\ \hline
	\end{tabularx}
\end{table}

\begin{table}[t]
	\centering
	\caption{Performance of fusion models on ActivityNet v1.3. Performance is measured by per-class mean average precision (mAP) and top-$3$ prediction accuracy.
		In ``Visual + Audio'' setting, we combine the visual and acoustic modeling system.
		On the testing set, we present the results of ``Final Ensemble'' where all components trained on training plus validation data are combined.}
	\vspace{5pt}
	\label{tb:fusion}
	\begin{tabularx}{1\linewidth}{|l *{2}{|Y}|}
		\hline
		Validation Set & mAp       	& Top-3 Acc. \\ \hline
		Visual         & $90.4\%$ 	& $95.2\%$  \\ \hline
		Audio          & $15.2\%$   & $29.1\%$          \\ \hline
		Visual + Audio & $90.9\%$ 	& $95.6\%$  \\ \hline\hline
		Testing Set    & mAP       	& Top-3 Acc. \\ \hline
		Visual CNN (Single)     & $91.2\%$ 	& $95.6\%$  \\ \hline
		Final Ensemble & $93.2\%$ 	& $96.4\%$  \\ \hline
	\end{tabularx}
\end{table}

We train our models on the official training set of ActivityNet v$1.3$ dataset~\cite{Caba2015anet}.
There are $10,024$ videos for training, enclosing $15410$ activity instances from $200$ activity classes.
The validation set contains $4926$ videos and $7654$ activity instances. We study the performance of our approach on this validation set.
The final testing set comprises $5044$ videos and is not annotated with any activity instance. We report the performance of our proposed models on this set according to the feedback of the test server of the challenge. 
Models for this setting are trained with the union of training and validation set.

In experiments, we compare the performance of temporal segment networks~\cite{Wang2016ECCV} using several network architectures, including
BN-Inception~\cite{IoffeS15}, Inception V3~\cite{Szegedy2015ICP2}, and ResNet~\cite{He2015ResNet}. The performance of different network structures for spatial and temporal stream are summarized in Table~\ref{tb:cnn}. 
To analyze the effect of different training strategies, we compare the performance of appearance modeling CNNs with these strategies. The results are presented in Table~\ref{tb:cnn_tricks}.
The contributions of appearance and motion CNNs are also summarized in Table~\ref{tb:visual}.
Then we report the performance of the two components in the acoustic analysis systems in Table \ref{tb:audio}. 

Finally, we evaluate the fusion of visual analysis system and audio analysis system on  both the validation and testing set.
The results are illustrated in Table~\ref{tb:fusion}.
The best mAP achieved by the final ensemble is $93.2\%$.
We also took one chance on the testing server to evaluate a combination of one appearance CNN and one motion CNN. Its results are presented as ``Visual CNN (Single)'' in Table~\ref{tb:fusion}. It is exciting to see using this ``single model'' setting we can still achieve a reasonable mAP of $ 91.2\% $, which may better fit for industrial applications.

\section{Conclusions}
This paper has proposed an action recognition method for classifying temporally untrimmed videos. 
It is based on the idea of combining visual analysis and acoustic analysis. 
The results show that by carefully designing the visual and acoustic analysis systems and combining them, 
we can achieve exciting results in video classification tasks and boost the performance of state-of-the-art methods.
Another fact to be noticed is that this high accuracy is achieved by evaluating only $ 1 $ frame per second, equivalent to only seeing around $4\%$ of all frames of input videos.
We believe this property is also very important for practically applying the system in industrial scenarios.

\section{Acknowledgment}
This work was supported by the \emph{Big Data Collaboration Research} grant from SenseTime Group (CUHK Agreement No. TS1610626) and ERC Advanced Grant Varcity (No. 273940).

{
\bibliographystyle{ieee}
\bibliography{deep}
}

\end{document}